# Real-world Transfer of Evolved Artificial Immune System Behaviours between Small and Large Scale Robotic Platforms

**Amanda M. Whitbrook · Uwe Aickelin · Jonathan M. Garibaldi**

**Abstract** In mobile robotics, a solid test for adaptation is the ability of a control system to function not only in a diverse number of physical environments, but also on a number of different robotic platforms. This paper demonstrates that a set of behaviours evolved in simulation on a miniature robot (epuck) can be transferred to a much larger-scale platform (Pioneer), both in simulation and in the real world. The chosen architecture uses artificial evolution of epuck behaviours to obtain a genetic sequence, which is then employed to seed an idiotypic, artificial immune system (AIS) on the Pioneers. Despite numerous hardware and software differences between the platforms, navigation and target-finding experiments show that the evolved behaviours transfer very well to the larger robot when the idiotypic AIS technique is used. In contrast, transferability is poor when reinforcement learning alone is used, which validates the adaptability of the chosen architecture.

**Keywords** Artificial Immune Systems (AIS) · Idiotypic Networks· Evolutionary Robotics· Cross Platform Transfer· Genetic Algorithms

# 1 Introduction

Evolutionary robotics is a field that is concerned with the genetic encoding of autonomous, robotic control systems and their improvement by artificial evolution. Ideally the end product should be a controller that has evolved rapidly and has the properties of robustness, scalability and adaptation. However, in practice it proves difficult to achieve all of these goals without introducing an additional mechanism for adaptability since behaviour is essentially an emergent property

Amanda M. Whitbrook · Uwe Aickelin · Jonathan M. Garibaldi
Intelligent Modelling and Analysis Research Group (IMA), School of Computer Science, University of Nottingham, Nottingham, NG8 1BB
Tel.: +44 115 8466568
Fax: +44 115 8467877
E-mail: amw, uxa, jmg@cs.nott.ac.uk

of interaction with the environment [1]. Thus, a major challenge facing evolutionary robotics is the development of solutions to the problem of brittleness via the design of controllers that are suited to dynamically changing environments.

The characteristics of the robot body and its sensorimotor system may be regarded as part of the environment [2] as all embodied systems are physically embedded within their ecological niche and have a dynamic reciprocal coupling to it [3]. Indeed, artificial evolution often produces control systems that rely heavily on body morphology and sensorimotor interaction [4], and when these are subsequently altered the changes can affect behavioural dynamics drastically. Thus, a solid test for robustness, scalability and adaptation is the ability of an evolved control system to function not only in different physical environments, but also on a number of robotic platforms that differ in size, morphology, sensor type and sensor response profile. This paper is therefore concerned with demonstrating the theoretical and practical cross platform transferability of an evolutionary architecture designed to combat adaptation problems.

Adaptation is usually made possible through the introduction of additional mechanisms that permit some kind of post-evolutionary behaviour modification. The architecture used here falls into the general category of approaches that combine evolution or long term learning (LTL) with a form of lifelong or short term learning (STL) to achieve this [5]. The particular technique consists of the rapid evolution of a number of diverse behaviour sets using a Webots [6] simulation (LTL) followed by the use of an idiotypic artificial immune system (AIS) for selecting appropriate evolved behaviours as the robot solves its task in real time (STL). The approach differs from most of the evolutionary schemes employed previously in the literature in that these are usually based on the evolution of neural controllers [7] rather than the actual behaviours themselves.

Previous papers have provided evidence that an idiotypic AIS architecture has advantages (including greater adaptability) over a reinforcement learning (RL) scheme when applied to mobile robot navigation problems [8], and have shown that the idiotypic LTL-STL architecture permits transference from the simulator to a number of different real-world environments [9] using small epuck robots, see Figure 1. The chief aim of this paper is, therefore, to supply further support for the robustness, scalability and adaptability of the architecture by showing that it can be extended to the much larger-scale Pioneer P3-DX robots, see Figure 2. For this purpose, the behaviours evolved on the epuck in the We-bots simulator are transplanted onto the Pioneer, which is then controlled by the idiotypic AIS algorithm. The test bed takes the form of Player's Stage [10] simulator with synthetic Pioneers, and a real-world laboratory environment with real Pioneer robots. The results successfully demonstrate the adaptability and scal-ability of the idiotypic methodology. Furthermore, control experiments, which utilize RL as the Pioneer control mechanism, provide strong empirical evidence that the idiotypic network is key to achieving the required levels of adaptability.

The remainder of this paper is structured as follows. Section 2 introduces some essential background information about the problem of platform transfer in mobile robotics and previous attempts to achieve it. The section shows how the solution can be broken down into the development of an adaptive control scheme to counter hardware differences, and the employment of a modular program structure to enable compatibility between heterogeneous software. In addition, software inconsistencies that arise out of using different hardware also need to be addressed. Section 3 describes the adaptive, idiotypic LTL-STL control system and the novel behaviour encoding that permits genetic diversity. Section 4 details the particular hardware and software differences between epucks and Pioneers and explains how these are handled within the cross platform transfer. Section 5 illustrates the test environments and experimental set-up and Section 6 presents and discusses the results. Section 7 concludes the paper.

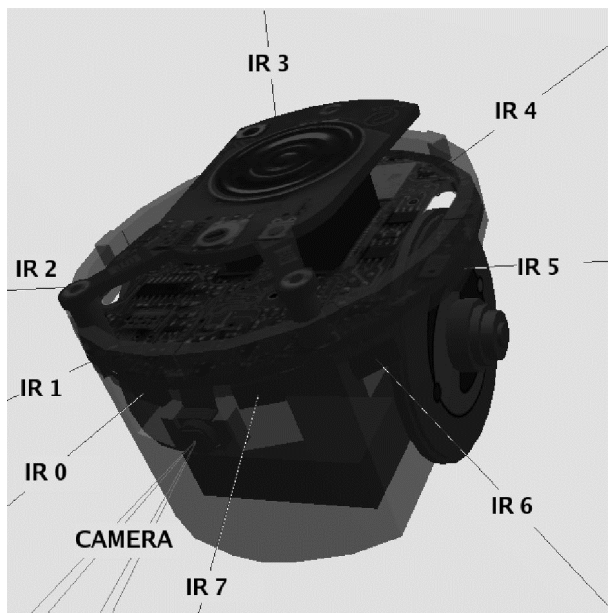

**Fig. 1 The epuck robot showing its infrared sensors and camera**

## 2 Background and Relevance

Cross platform transfer of an intelligent robot-control algorithm is a highly desirable property since more generic software is more marketable for vendors and more practical for users with more than one robot type. Furthermore, software that is robust to changes in a user's hardware requirements is particularly attractive. However, transferability between platforms is difficult to achieve and is hence extremely rare in mobile robotics [2]. This is primarily due to hardware differences such as size, morphology, mechanical structure, and sensor and actuator types and characteristics. In addition, the spatial relationships between the body, actuators and sensors may also impede transfer. Such hardware variations constitute a drastic change of the environment from an ecological perspective, and any control algorithm intended for use with such a variety of hardware configurations would need to be highly adaptable. Furthermore, the software required for controlling hardware is usually very much dependent on the type, which leads to further complications, and since modern mobile-robot systems are distributed systems, platform software differences such as diversity of middleware, operating systems, communications protocols, and programming languages and their libraries [11] also present practical challenges when designing multi-platform software. Hence, as well as being adaptable, control algorithms also need a high level of systems engineering to ensure compatibility.

Despite its rarity, platform transfer for evolved control systems is reported in the literature. Floreano and Mondada [2,4] use an incremental approach to evolve artificial neural networks for solving a looping maze navigation problem. Evolution begins with a real miniature Khepera robot and gradually moves to a Koala, a larger, more fragile robot supplied by the same manufacturer. Within this architecture, previously evolved networks are gradually adapted, combined, and extended to accommodate the changing morphology and sensorimotor interfaces [4]. However, the scheme possesses some significant drawbacks. The use of physical robots for the evolution is highly impractical due to the excessive time and resources required. For example, adaptability to the Koala platform emerges only after 106 generations on the real Khepera and an additional 30 on the Koala (each generation taking approximately 40 minutes). Also, if each new environment or platform requires additional evolution, then there is no controller that is immediately suitable for an unseen one.

Another consideration is that the Koala was deliberately designed to support transfers from the Khepera [2]. These design similarities facilitate compatibility between software, but there are also hardware similarities, which place limits on the analogy of different hardware representing different environments. In particular, although the two robots have different sizes and shapes, they use the same distance sensor type and camera, and the wheel rotation configuration of the Koala is very similar to that of the Khepera.

Floreano and Urzelai [12,13] also evolve a neural network to control a light-switching robot, but evolve the mechanisms for self-organization of the synaptic weights rather than the weights themselves. This means that the robot is rapidly able to adapt its connection weights continuously and autonomously to achieve its goal. The authors transfer the evolved control system from simulation to a real Khepera and also from a simulated Khepera to a real Koala. They report no reduction in performance following the transfer. However, the task is very simple and the environment constructed for the robot is sparse, requiring minimal navigational and obstacle avoidance skills.

The research presented here uses more complex tasks and environments to demonstrate that behaviours evolved on a simulated epuck can be used by a larger, unrelated robot that has not deliberately been designed for ease of transfer (the Pioneer P3-DX). This represents a more difficult platform transfer exercise than has been attempted before, both in terms of the requirement for a more adaptable algorithm (there are more hardware differences) and the need for a system design that supports both platforms (the software differences are much more complex). As with Kheperas and Koalas, Pioneer and epuck robots differ from each other in mechanical structure, body size (see Figure 2), body shape and wheel size, but they also have a different wheel rotation configuration and use a different type of distance sensor and camera. The Pioneer uses sixteen linear sonar sensors with very different characteristics to the eight nonlinear IR sensors of the epuck, and these also have a different spatial arrangement, see Figure 3. These differences place an additional burden on the control software, i.e., there is a demand for greater adaptability than that demonstrated by the systems in [2], [4], [12] or [13]. In addition, the Pioneer is produced by a different manufacturer and uses different middleware, programming language, libraries and simulator (Stage [10]), which complicates the transfer in terms of compatible software requirements. A full comparison between the two platforms is provided in Section 4, which highlights the particular hardware and software differences (Table 2), and discusses them in more depth.

It is worth mentioning that transfer of a control system between epucks and Pioneers does not only suggest the adaptability of the control algorithm in question, it is also of practical value. Pioneer behaviours cannot be evolved directly on the Stage simulator within a realistic time frame as Stage is not fast or accurate enough, and control systems used in the Webots programming environment are not directly transferable to real Pioneers. Moreover, simulation of the epuck in Webots requires a 3D model that is readily available, so it is computationally much cheaper to reuse the epuck's evolved behaviours in Stage rather than to design a complex 3D Pioneer model for Webots.

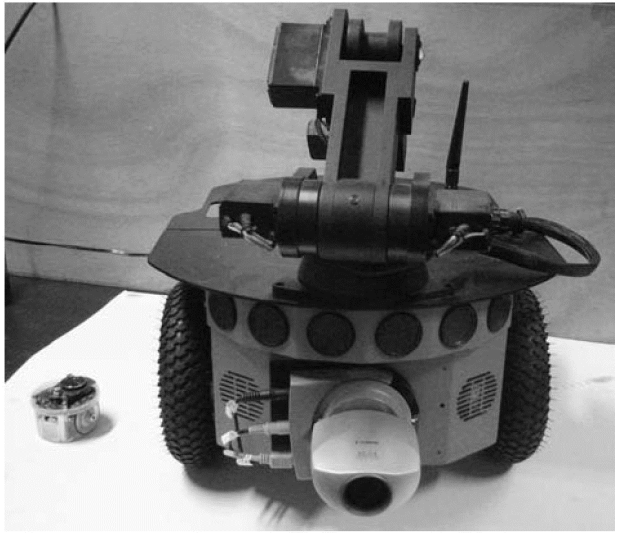

**Fig. 2 Showing the different sizes of the epuck (left) and Pioneer (right) robots. Six of the Pioneer's front sonar can be seen as large discs around the circumference, just above the protruding camera**

The aim of this paper is to show adaptability of the LTL-STL idiotypic AIS control algorithm by demonstrating that behaviours can be evolved on the epuck and transferred to the Pioneer when the idiotypic mechanism is turned on. For this purpose, an RL version of the algorithm is also used to highlight the difficulties with the platform transfer. However, a logical extension of this research is the development of a more generic architecture that would allow diverse behaviours to evolve rapidly on a simple, virtual robot, possibly even a fictitious one, but permit the use of such behaviours on a number of vastly different and more complex robotic platforms operating in the real world. In the future this may lead to standardized methods of control system generation, making it possible to dispense with the need for designing complex simulation models. This paper hence represents a first step towards achieving this goal.

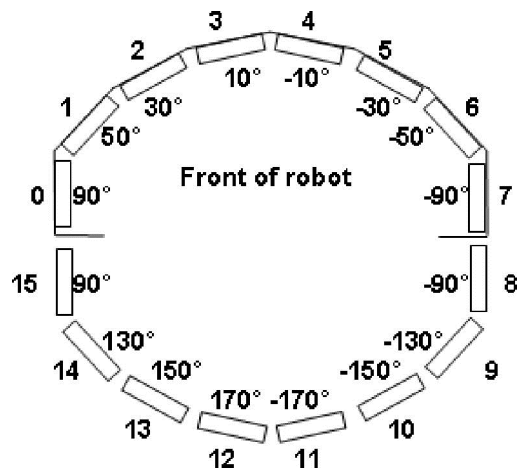

**Fig. 3** The spatial arrangement of the sonar sensors on the Pioneer robot

## 3 System Architecture

3.1 Artificial Immune Systems and the Behavioural Encoding

In the vertebrate immune system B cells secrete antibodies, which help to neutralize foreign objects (termed antigens) such as bacteria and viruses. When neutralization occurs, a region of the antibody known as the paratope binds to a region of the antigen known as the epitope and the presence of the antigen stimulates an increase in the concentration of antibodies with paratopes that match the presented epitope. However, antibodies also possess an epitope region known as the idiotope and can bind to each other. According to Jerne's idiotypic network theory [14], this facilitates further antibody stimulation and suppression, so that the concentrations are governed by global network interactions. The theory views the immune response as a distributed system, with each individual component contributing to the final outcome, which shows emergent behaviour. Hence, complex dynamics arise out of a multiplicity of relatively simple interactions [15].

AIS systems in general mimic the properties of the immune system, and idio-typic AIS algorithms are often based on Farmer *et al.*'s computational model [16] of Jerne's theory, where antigens are analogous to the environmental information as perceived by the sensors and antibodies are analogous to the behaviours of the robot. The Farmer model is a popular choice for STL in robotic control systems [17] since it encapsulates the emergent properties of the idiotypic system, which permits greater flexibility for determining a robot's actions. For this reason, Farmer *et al.*'s idiotypic model is used for the STL phase of the control architecture described here, as the STL phase requires a high level of adaptability. The LTL phase is provided by a GA, which should increase STL adaptability by evolving very diverse behaviour sets for the idiotypic AIS to choose from.

**Table 1** System antigens and basic antibody (or behaviour) types

| System Antigens | | Basic Antibody Types | |
|---|---|---|---|
| No. | Description | Type | Description |
| 1 | Target unseen | 1 | Wandering using either a left or right turn |
| 2 | Target seen | 2 | Wandering using both left and right turns |
| 3 | Obstacle right | 3 | Turning forwards |
| 4 | Obstacle rear | 4 | Turning on the spot |
| 5 | Obstacle left | 5 | Turning backwards |
| 6 | Collision right | 6 | Tracking targets |
| 7 | Collision rear | | |
| 8 | Collision left | | |

Here, eight antigens (numbered 1 - 8) are identified based on the robot's possession of distance-measuring sensors (IR or sonar) and a camera for tracking coloured objects, see Table 1. In addition, six types of fundamental antibody or behaviour types are used, see Table 1. More detailed individual behaviours are described by combining the behaviour type $T$ with additional attributes and assigning values to them. The additional attributes are speed $S$ in epuck speed units per second ($\psi$ per second), frequency of turn $F$ (% of time), angle of turn $A$ (% reduction in one wheel speed), direction of turn $D$ (either 1 - left or 2 - right), frequency of right turn $Rf$ (% of time) and angle of right turn $R_a$ (% reduction in right wheel speed). This structure means that, potentially, a vast number of diverse behaviours can be created. However, since the LTL phase of the architecture involves a GA, it makes sense to impose limits to the attribute values [18] in order to strike a balance between reducing the size of the search space, which increases speed of convergence, and maintaining diversity. More details on the behavioural encoding are provided in Section 4.

3.2 LTL Phase

The LTL phase consists of an RL-assisted genetic algorithm (GA) that evolves a suitable behaviour for each antigen. An integer-coded GA is used rather than a binary one as this permits better optimization, a faster search and easier integration with other approaches [19]. The GA works by selecting two different parent robots via the roulette-wheel method and determining behaviour attribute values for their offspring. Complete antibody replacement occurs according to a prescribed mutation rate $\gamma$; when this occurs, a new random behaviour is assigned to the child robot for the particular antigen, i.e., both the parent behaviours are ignored.

When there is no complete replacement, parent attribute values are distributed to the offspring by a number of means. If the parent behaviour types *T* are different then it makes no sense to use crossover, so the child adopts the complete set of attribute values of one parent only, which is selected at random. If the types are the same, then crossover can occur, either:

− by taking the averages of the two parent values,
− by randomly selecting a parent value, or
− by taking an equal number from each parent according to set patterns.

In each case, the crossover method is determined randomly with equal probability. The purpose behind this approach is to generate as much diversity as possible both generally and particularly when offspring are produced by the same two parents; if the same crossover method were to be used each time, then diversity would not be as high. (Note that natural reproduction uses recessive genes for this purpose.) In addition, mutation of a single attribute value, i.e. an increase or decrease by between 20% and 50% of the original value may also take place according to the mutation rate $\gamma$, provided that complete replacement has not already occurred.

The RL component of the GA constantly assesses the performance of the behaviours during evolution so that poorly-matched ones are replaced with newly-created ones when the need arises, which accelerates the GA. The combination represents an iterative hybrid technique and, as such, has several advantages for online problem solving. The GA component produces a reliable solution that benefits from a stochastic, global search of the solution space and is capable of escaping from sub-optimal points during optimization. However, GAs can demonstrate inaccuracies and are often slow [20] since they use minimal *a priori* knowledge and do not exploit local information. Here, the RL component effectively balances the trade-off by working at the local level. This maintains the reliability of the GA whilst decreasing computation time and improving accuracy [19]; the hybrid technique thus takes advantage of the exploitative properties of the reinforcement algorithm and the explorative properties of the GA.

All the test problems are assessed by measuring both the primary objective (a fast completion time $t_i$, $i = 1, ..., x$) and level of constraint violation or cost (number of collisions $c_i$, $i = 1, ..., x$), see Section 5 for details. Thus, the relative fitness $\mu_i$ of each member of the population is calculated using:

$$\mu_i = \frac{1}{t_i + \rho c_i} \left( \sum_{k=1}^{x} (t_k + \rho c_k) \right)^{-1}, \qquad (1)$$

where $\rho$ represents the weighting given to the cost function ($\rho = 1$ here) and *x* is the number of robots in the population. After convergence, the fittest robot in the final population is selected. However, since the idiotypic network requires a number *n* of distinct behaviours for each antigen, the whole process is repeated *n* times in order to obtain *n* robots from separate populations that never interbreed. This is an alternative to selecting a number of robots from a single, final population and means that greater diversity can be achieved with smaller population sizes without increasing convergence time, see [18]. The attribute values representing the behaviours of the $n$ robots and their final reinforcement scores are saved as a genetic sequence (a simple text file) for seeding the AIS system. There are thus $n$ solution sets, each consisting of $y$ antibodies or behaviours. Note that $y$ represents both the number of antigens (eight) and the number of antibodies in each solution set, since there is one antibody for each antigen.

3.3 STL Phase

The AIS system reads the genetic sequence generated by the LTL phase ($n = 5$ as recommended in [18]), and then calculates the relative fitness $\mu_i$ of each solution set $i = 1, ..., n$ using (1), where $\rho = 8$ to increase the weighting given to the cost function. It then produces an $n \times y$ matrix $P$ (analogous to an antibody paratope) representing the reinforcement scores, or degree of match between antibody paratopes and antigen epitopes. The elements of this matrix ($P_{ij}$ $i = 1, ..., n, j = 1, ..., y$) are calculated by multiplying each antibody's final reinforcement score by the relative fitness $\mu_i$ of the solution set that it belongs to. An $n \times y$ matrix $I$ (analogous to an antibody idiotope) is also created by assigning a value of 1.0 to the element corresponding to the minimum $P_{ij}$ for each $j$, and designating a value of 0.0 to all other elements. $I$ is used to drive the idiotypic network, i.e., it adjusts the degree of match of each antibody to the presented antigen by taking account of inter-antibody stimulation and suppression, which may result in a different antibody being selected. The matrix $P$ is adjusted after every iteration through RL, but $I$ remains fixed throughout as previous work has shown that parameter selection is difficult for a stochastically variable idiotope because of its unpredictability, see Section 5.1.

The antibody with the highest degree of match to the presented antigen $m$, i.e. the antibody with the highest paratope value $P_{im}$, $i = 1, ..., n$, becomes the antigenic antibody. Once this is established, the idiotypic network suppresses dissimilar antibodies, and stimulates similar ones by comparing the idiotope of the antigenic antibody with the paratopes of the other antibodies to determine how much each is stimulated, and by comparing the antigenic paratope with the idiotopes of the others to calculate how much each should be suppressed. Further explanation of the stimulation and suppression mechanism is presented in [8]. If the antigenic antibody is the $l$th, and $y$ represents the number of antigens, equations (2) and (3) govern the increase $a$ and decrease $\delta$ in affinity to $m$ for each of the $n$ solution sets when stimulation and suppression occur respectively:

$$\alpha_{im} = k_1 \sum_{j=1}^{y} (1 - P_{ij}) I_{lj} C_{ij} C_{lj} \quad i = 1, ..., n, \quad (2)$$

$$\delta_{im} = k_2 \sum_{j=1}^{y} P_{lj} I_{ij} C_{ij} C_{lj} \quad i = 1, ..., n. \quad (3)$$

Here, $k_1$ and $k_2$ are constants that can be used to control the relative effects of stimulation and suppression and $C$ represents concentrations of antibodies. If ($P_{im}$)$_1$ represents the existing paratope value or degree of match to the epitope of $m$, then a new affinity ($P_{im}$)$_2$, which accounts for idiotypic effects, is given by:

$$(P_{im})_2 = (P_{im})_1 + im - \delta_{im}. \qquad (4)$$

However, the selected antibody is not that with the highest new affinity value; it is the matching antibody with the highest concentration. Each antibody begins with a number of clones $N_{t0} = 1000$ (the initial value is based on recommendations in [9]), which fluctuates with time according to a variation of Farmer's equation [16]:

$$(N_{im})_{t+1} = b(P_{im})_2 + (N_{im})_t(1 - k_3), \qquad (5)$$

where ($N_{im}$)$_t$ represents the number of clones of each antibody matching the invading antigen $m$, ($P_{im}$)$_2$ is the newly-adjusted paratope value of each of these antibodies (representing the new strength of match to $m$), $b$ is a scaling constant and $k_3$ is an antibody death rate constant. The concentration $c_{ij}$ of each antibody in the system consequently changes according to:

$$c_{ij} = \frac{\phi N_{ij}}{\sum_{p=1}^{n} \sum_{q=1}^{n} N_{pq}}, \qquad (6)$$

where $\phi$ is another scaling factor that can be used to control the levels of inter-antibody stimulation and suppression. The chosen antibody is the one with the highest concentration value, selected from all those that match antigen $m$, i.e. that with the maximum $c_{im}$. It may be the same as the antigenic antibody, but it may be some other that matches the presented antigen, in which case an *idiotypic difference* is said to occur. It is worth noting that the algorithm merely records the number of clones, it does not have to store thousands of instances of a complex class structure, so there is no additional computational burden in utilizing the concentration metric.

## 4 Mechanisms of Cross Platform Transfer

In order to create a control system that facilitates a successful cross platform transfer, three top-level considerations are necessary:

1. **Adaptability** - A highly adaptive control system must be used, i.e., one that is robust to the equivalent of an environmental change brought about by use of different hardware. Here, this aspect is dealt with by incorporating an idiotypic network with high behavioural diversity into the architecture, see Section 3.
2. **Inconsistency Handling** - There is a need to examine the control program parameters and other algorithmic inconsistencies that arise out of using different hardware. For example, the perceived environmental information may be heavily dependent upon parameters governed by sensor characteristics.

**Table 2** Differences between the Pioneer and epuck robotic platforms. In general, software differences demand systems engineering to achieve compatibility, but hardware differences require adaptability and inconsistency handling.

| Attribute | Pioneer P3-DX | epuck | Difference | Items |
|---|---|---|---|---|
| Manufacturer | MobileRobots Inc | EPFL | – | – |
| Simulator | Stage | Webots | – | – |
| Operating system | Linux | N/A | – | – |
| Middleware | Player | Webots | Software | 3 |
| Blob finding software | Player | Weblobs | Software | 3 |
| Communications protocol | Wireless TCP/IP | Bluetooth | Both | 1, 2, 3 |
| Wheel radius (cm) | 9.50 | 2.05 | Hardware | 1, 2 |
| Wheel width (cm) | 5.00 | 0.20 | Hardware | 1, 2 |
| Axel length (cm) | 33.00 | 5.20 | Hardware | 1, 2 |
| Body material | Aluminium | Plastic | Hardware | 1, 2 |
| Body length (cm) | 44 | 7 | Hardware | 1, 2 |
| Body width (cm) | 38 | 7 | Hardware | 1, 2 |
| Body height (cm) | 22 | 4.8 | Hardware | 1, 2 |
| Weight (kg) | 9 | 0.15 | Hardware | 1, 2 |
| Body shape | Octagonal | Circular | Hardware | 1, 2 |
| Sensor type | Sonar | Infrared | Hardware | 1, 2 |
| No. of sensors | 16 | 8 | Hardware | 1, 2 |
| Sensor range | 15cm to 5m | 0 to 6cm | Hardware | 1, 2 |
| Camera | Canon VC-C4 | VGA | Hardware | 1 |

3. **Systems Engineering** - The control program must be designed so that it is robust to actual platform software differences, for example the Pioneer's use of Player's libplayerc++ library, which specifies velocity in a different way to the Webots framework used by the epucks.

Table 2 summarizes the Pioneer and epuck platform differences and categorizes them into hardware and software classes of problem. It also highlights which considerations (1 to 3 above) are applicable in each case.

One of the main hardware differences between the platforms is the use of sonar sensors on the Pioneer as opposed to infrared on the epuck (see Figure 1), and their different spatial arrangement (see Figure 3), which imposes a greater demand for control system adaptability (item 1) than has been required for previous cross platform transfer exercises, for example [2], [4], [12] and [13]. It also introduces a need for constructing alternative methods for reading the sonar and translating their data into antigen codes (item 2). In this architecture, the antigens indexed 3 to 8 describe an obstacle's orientation with respect to the robot (right, left or rear) and classify its distance from the robot as either "obstacle" (avoidance is needed) or "collision" (escape is needed). Thus, two threshold values $\tau_1$ and $\tau_2$ are required to mark the boundaries between "no obstacle" and "obstacle" and between "obstacle" and "collision" respectively. The epuck's IR sensors are nonlinear and correspond to the quantity of reflected light, so higher readings mean closer obstacles. In contrast, the Pioneer's sonar readings are linear, denoting the estimated proximity of an obstacle in metres, so lower readings mean closer obstacles. Since direct conversion is difficult, the threshold values $\tau_1$ and $\tau_2$ (250 and 2400 for the epuck) are determined for the simulated and real Pioneer by empirical observation of navigation through cluttered environments ($I_1 = 0.15$, $I_2 = 0.04$ in simulation and $I_1 = 0.30$, $I_2 = 0.04$ in the real world).

Additionally, in order to determine the orientation of any detected obstacle, the epuck uses the index of the maximum IR reading, where indices 0, 1 and 2 correspond to the right, 3 and 4 correspond to the rear and 5, 6 and 7 correspond to the left, see Figure 1. For the Pioneer it is necessary to use the index of the minimum sonar reading and encode positions 4 to 9 as the right, 10 to 13 as the rear and positions 0 to 3 and 14 to 15 as the left, due to the different spatial arrangement of the sensors, see Figure 3.

The chief software consideration is the use of different application programming interfaces (APIs) for the Pioneer and epuck robots (item 3). This affects how velocity is expressed, how blob finding is implemented, how the different sensor and actuator components are instantiated and called and how they pass data to the main control module. The velocity problem is discussed next, and the modular structure of the control architecture, which provides a solution to most of the API incompatibility problems, including implementation of blob finding, is presented in Section 4.1.

Following convergence of the GA, the selected antibodies or behaviours are encoded as nine integers in a simple text file that contains all the genetic information necessary to reproduce them. The first integer represents the antigen number, and the next seven represent the behavioural attributes $T$, $S$, $F$, $A$, $D$, $R_f$ and $R_a$. The last integer is the final reinforcement score attained by the behaviour prior to convergence. The genetic sequence encodes the principal wheel speeds in epuck speed units per second ($\psi$ per second) where $\psi = 0.00683$ radians; a speed value of 600 $\psi$ per second thus corresponds to $600 \times 0.00683 = 4.098$ radians per second. An example line from a genetic text file is: 0 2 537 80 51 2 37 76 50, which encodes wandering in both directions with a speed of 537 $\psi$ per second, turning 80% of the time and moving directly forwards 20% of the time. The robot turns right 37% of this time (i.e. 29.6% of the total time) by reducing the speed of the right wheel by 76%, and turns left 63% of this time (i.e. 50.4% of the total time) by reducing the speed of the left wheel by 51%. Note that epuck robots must turn by generating differential velocity between the two wheels, as the platform uses skid steering (fixed wheels) rather than Ackermann steering (freely turnable wheels). A particular genetic sequence thus governs how the left and right wheel speeds change with time.

In theory, the behavioural encoding may be extended to any skid-steering, non-holonomic, mobile-robot, since the wheel motions of such robots are fully described by their changing speeds. Furthermore, since the output from the LTL phase is a simple text file, any program is capable of reading it and extracting the information necessary to form the wheel motions. Moreover, specification of the speeds in radians per second permits automatic scaling between different-sized environments, without requiring knowledge of the particular scales involved, since wheel size is generally related to the scale of the environment.

As Pioneers are skid-steer robots, use of the genetic sequence coupled with a simple conversion of $\psi$ per second to radians per second, as described above, would be adequate to cater for the scaling differences if the two platforms did not use different APIs. However, the epuck is programmed using the Webots

C/C++ Controller API, where robot wheel speeds are set using the differen-tial wheels set speed method, which requires the left and right wheel speeds in $\psi$ per second as its arguments. In contrast, the Pioneer robot is programmed using libplayerc++, a C++ client library for the Player server [21]. In this library, the angular and linear components of the robot's velocity are set separately using yaw $\omega$ and velocity $v$ arguments for the SetSpeed method of the Position2dProxy class, where $\omega$ is expressed in radians per second and $v$ is expressed in metres per second. As methods for encoding the genetic sequence into left $L$ and right $R$ epuck wheel speeds already exist, it is computationally cheaper to reuse these methods on the Pioneer and simply convert them into equivalent $\omega$ and $v$ arguments. The conversion essentially represents a program design change (item 3) and is given by:

$$v = \frac{\psi r_p (R + L)}{2}, \quad (7)$$

$$\omega = \frac{\zeta \psi r_e (R - L)}{a_e}, \quad (8)$$

where $r_p$ is the radius of the Pioneer wheel, $r_e$ is the radius of the epuck wheel, and $a_e$ is the axle length of the epuck. However, the parameter $\zeta$ (determined by repeated trial and error) represents an inconsistency handling change (item 2) and is introduced so that the Pioneer can replicate the angular movement of the epuck more accurately. In these experiments, parameter values of 1.575 and 0.750 are used in the simulated and real worlds respectively, which produce approximately the same angle of turn per unit wheel speed difference as the epuck. The inconsistency may be caused by the different communications proto-col, body weights, dimensions and materials in the epuck and Pioneer, but it is likely that a smaller $\zeta$ value is necessary for the Pioneer in the real world because factors such as slippage of the wheels and other physical forces are unaccounted for in the simulations; i.e., the real Pioneer tends to turn more rapidly than its virtual counterpart for a given wheel speed difference.

4.1 Modular Control Structure

For the STL phase, the aim is to have a main program that can be used for con-trolling both the Pioneer and epuck robots. This should call separate modules, some of which are robot-specific for handling hardware initialization and control. The STL control program is thus split into generic *main* and *behaviour* modules and robot-specific *robot* and *blobfinder* modules. The pseudo code shown below illustrates the modular structure; each block shows the module it calls and the method it uses within that module. Blocks marked with an asterisk are dealt with in the main program and do not call other modules.

```
1           Initialize robot (ROBOT --> InitializeRobot() --> InitializeSensors())
2           Read genetic sequence *
3           Build matrices P and I *
            REPEAT
4               Read sensors (ROBOT --> ReadSensors ( ) )
5               Read camera (BLOBFINDER --> GetBlobInfo ( ) )
```

```
6       Determine antigen code *
7       Score previous behaviour using reinforcement learning *
8       Update P *
9       Select behaviour *
10      Update antibody concentrations *
11      Execute behaviour (BEHAVIOUR--> Execute ( ))
   UNTIL stopping criteria met
```

The modular structure allows most of the inconsistency handling to be addressed within robot-specific modules. Thus, the only blocks of the pseudo code that require changes for the Pioneer platform are 1, 4, and 5. Since these are dealt with by calling other modules, the main program can be wholly reused, although an additional two lines in block 11 are necessary to convert the wheel speeds to the libplayerc++ format.

Separate blob finding modules are necessary since the Pioneer robot is able to use methods belonging to the BlobfinderProxy class of libplayerc++, see [21], but the Webots C/C++ Controller API has no native blob finding methods and hence has to use its own software (Weblobs, which was developed as part of this research). The objective of blob finding is to determine whether blobs (of the target colour) are visible, and if so, to establish the direction (left, centre or right) of the largest from the centre of the field of view. The blob finding modules implement very similar algorithms in this respect, so that the two robot platforms collect the same information, i.e. the identity of the largest blob and its direction. Tracking and following the blob is handled within the *behaviour* module so that there is no discrepency between the two platforms.

## 5 Test Environments and Experimental Set-up

The genetic behaviour sequences are evolved using 3D virtual epucks in the Webots simulator, where the robot is required to track blue markers in order to navigate through a number of rooms in a maze to the finish line, see Figure 4 and [18]. Throughout evolution, five separate populations of ten robots are used and the mutation rate $\gamma$ is set at 5%, as recommended by [18], for a good balance between maximizing diversity and minimizing convergence time. This permits the behaviours to evolve in approximately 20 minutes.

Following the LTL phase, the evolved behaviour sequences are used with 2D virtual Pioneer robots in three different Stage worlds named $W_1$, $W_2$, and $W_3$ (see Figure 5, Figure 6 and Figure 7 respectively), and with a real Pioneer in a real-world laboratory environment named $W_4$, see Figure 8. Note that the 3D world used for evolution is not related to any of the STL phase environments, i.e. the Stage worlds are not 2D representations of it, and it is not based on real world $W_4$. $W_1$ and $W_2$ require maze navigation and the tracking of coloured door markers, whilst $W_3$ and $W_4$ involve search and retrieval of a coloured block while navigating around other obstacles. In the real world the obstacles are the wall of the pen containing the robot and a number of cardboard boxes of varying sizes and shapes.

Sixty runs are performed in each world, thirty using the idiotypic selection mechanism, and thirty relying on RL only, i.e. 240 runs are performed in total. In addition, in $W_3$ and $W_4$ the obstacle positions, target location, and robot start point are changed following each pair of idiotypic and RL-only tests, so that the two control systems are effectively compared under many different scenarios, each representing a variation in the environment.

For all runs, the task time $t$ and number of collisions $c$ are recorded. However, in the simulator collisions are difficult to judge visually, and so these are registered each time any of the antigens 6, 7 or 8 are presented. In the real world collisions are counted when the robot physically impacts against a wall or an obstacle, which represents a more realistic measure of an actual collision. Additionally, for each task, robots taking longer than 900 seconds are counted as having failed and are stopped.

In order to demonstrate adaptability of the idiotypic LTL-STL architecture (i.e. transferability between platforms) and show that the idiotypic network component facilitates the adaptation, three hypotheses are formed:

– H1: The behaviours evolved on the simulated epuck transfer well to the simulated and real Pioneer platforms provided an idiotypic network is used to govern behaviour selection. This criteria can be measured by assessing task completion time, numbers of collisions and total number of failures for the idiotypic robots.
– H2: The behaviours evolved on the simulated epuck transfer poorly to the simulated and real Pioneer platform when an idiotypic network is not used to govern behaviour selection. Assessment is the same as for $H_1$, but for RL-only robots.
– H3: The idiotypic robots perform significantly better than the RL-only ones when the evolved epuck behaviours are transplanted onto them. Statistical tests to compare the results (task time and collisions) for the idiotypic and RL-only versions of the algorithm can be performed to measure the strength of this claim, see Section 6.

5.1 Parameter Selection

Previous work [8] has presented a rigorous analysis of the results of varying the parameters $b$ and $k_1$ for a virtual robot navigating in a two Stage worlds. This work shows that $b$ is stable (i.e. produces a significantly higher level of adaptability) within the region $40 < b < 160$, and that $k_1$ is less robust to change, producing stable solutions only between about $0.55 < k_1 < 0.65$. However, since the actual parameter values may be both problem and architecture dependent, it is much more meaningful to take account of the idiotypic difference rate $e$ (see Section 3.3) obtained when stable values are used. For the navigation tasks described in [8] a $e$ of about 20% is obtained in the stable region. Reference [9] is also able to demonstrate a significantly higher level of idiotypic adaptability when the mean $e$ value is about 20%, even though it uses different robots, a slightly different architecture, different control parameters, and different test problems (although $e$ showed a much higher standard deviation than in [8] for a given set of parameters). Prior work thus shows that $e$ is the important parameter for achieving idiotypic stability and that as long as other parameter choices

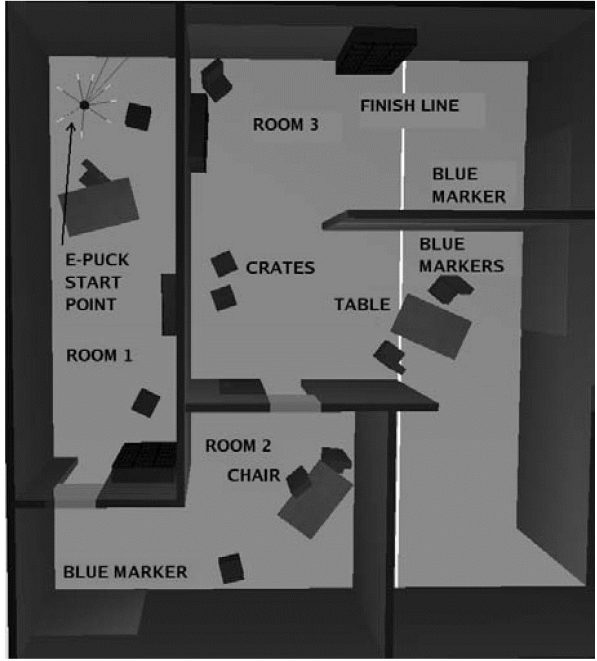

**Fig. 4 3D Webots world used in the epuck LTL phase**

combine to produce a suitable idiotypic difference rate, adaptable solutions will follow.

However, with all of the test problems described here a mean $e$ value of 20% proves difficult to achieve by manipulation of the parameters $k_1$, $k_2$, $b$ and , and as with [9], $e$ shows a high standard deviation for a given set of parameters. In addition, a series of preliminary experiments demonstrates that $e$ values less than about 35% do not appear to work well, and that, for these problems at least, high $e$ values of about 75±10% produce better results. A set of parameters that proves able to generate mean $e$ values in this region is thus selected for all the idiotypic experiments, and kept constant throughout. The parameters chosen are $b = 200$, $k_1 = 0.30$, $k_2 = 1.85$, and = 25. In addition, the parameter $k_3$ is set at 0 (as in [8] and [9]) as a fixed antibody death rate is not required.

## 6 Results and Discussion

In each world $t$ and $c$ show non-normality in their distributions (using the Shapiro-Wilks test) for both the idiotypic and RL-only robots. The Mann-Whitney U test (appropriate for non-normally distributed data) is hence used to test the differences between these control systems, with differences accepted as significant at the $p = 0.05$ level. Table 3 below shows median and interquar-tile range data for $t$ and $c$ in each world along with the number of fails, and Table 4 shows the $p$ values obtained in the Mann-Whitney tests. In addition,

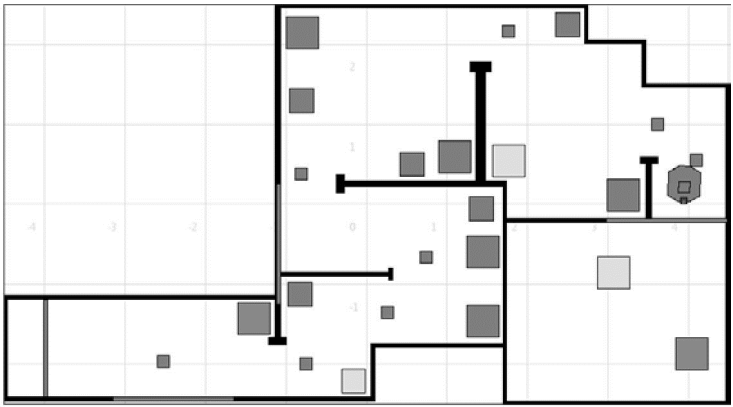

**Fig. 5** 2D Stage world W1 used in the Pioneer STL phase

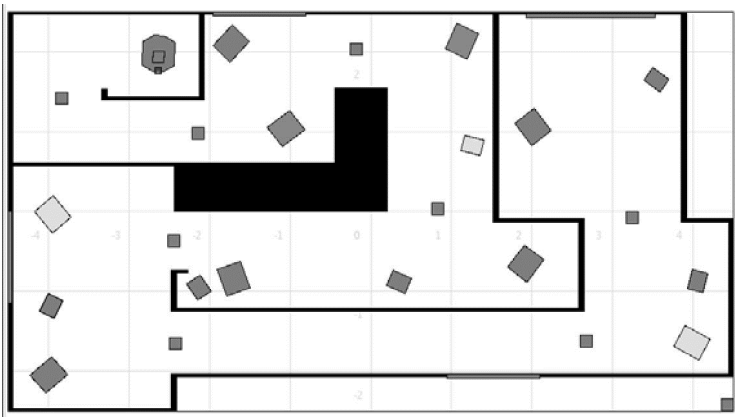

**Fig. 6** 2D Stage world W2 used in the Pioneer STL phase

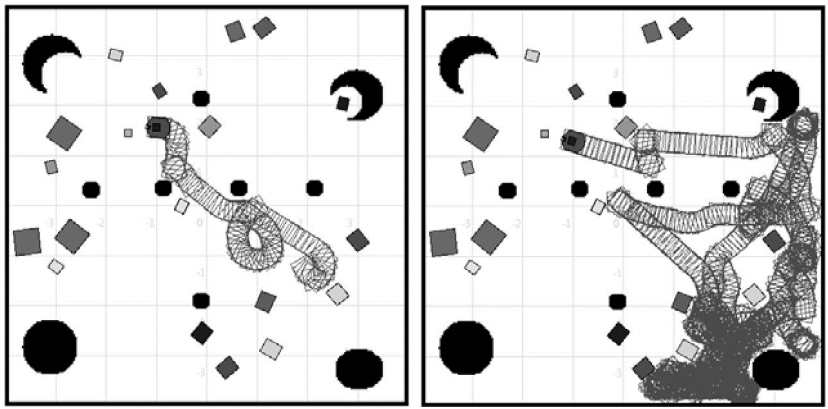

**Fig. 7** 2D Stage World W3 showing the trail of an idiotypic (left) and RL-only (right) robot

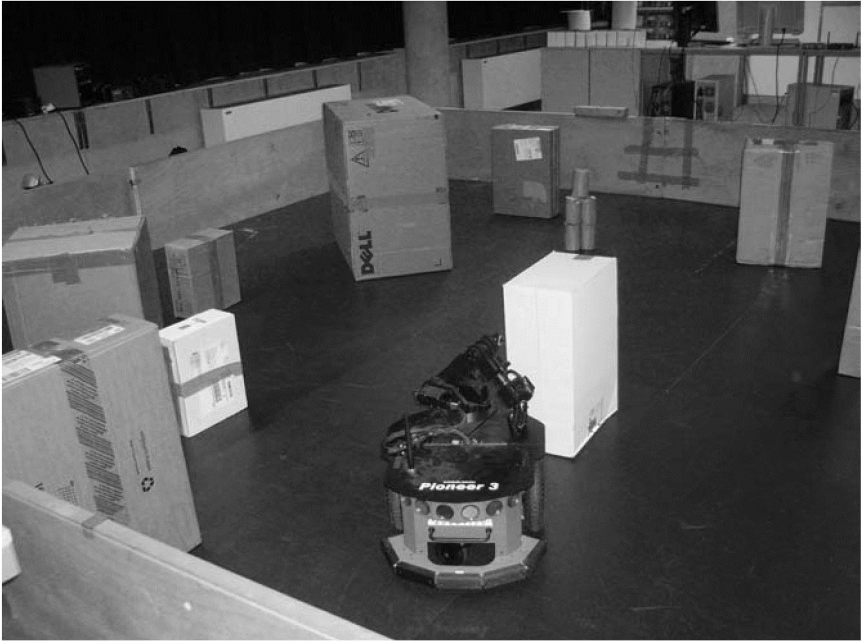

**Fig. 8** Real world $W_4$ and the Pioneer robot

**Table 3** Results of experiments using idiotypic and RL-only systems. F = % fail rate, IQR = interquartile range, med = median.

| World | Idiotypic Robots | | | | | RL-only Robots | | | | |
|---|---|---|---|---|---|---|---|---|---|---|
| | time (s) | | collisions | | F | time (s) | | collisions | | F |
| | med | IQR | med | IQR | | med | IQR | med | IQR | |
| $W_1$ | 148 | 109 | 1.0 | 2.0 | 0 | 180 | 277 | 2.0 | 2.0 | 17 |
| $W_2$ | 299 | 120 | 2.0 | 2.0 | 0 | 469 | 363 | 2.5 | 6.0 | 17 |
| $W_3$ | 117 | 173 | 1.0 | 1.0 | 0 | 361 | 475 | 1.0 | 2.0 | 7 |
| $W_4$ | 174 | 81 | 2.0 | 4.0 | 0 | 236 | 555 | 6.0 | 9.0 | 4 |

Table 4 also provides $A$ values, which are the result of scientific significance tests or $A$-tests designed to show the effects of the differences between the medians, i.e. whether they are scientifically important. Note that in this context an $A$ value of 0.50 indicates no effect but as values rise from 0.50 this shows increasing importance, with values above about 0.63 representing medium effects and values above about 0.70 representing large effects. The box-and-whisker plots in Figure 9 and Figure 10 display median, upper quartile, lower quartile, minimum and maximum data for time and collisions in the real world $W_4$.

Table 3 shows that when the idiotypic system is employed, the Pioneer robots prove able to navigate safely (the median and interquartile range of the number of collisions is very low), and they solve their tasks within the alloted time in all of the worlds, i.e. there is a 100% success rate in solving the problem, which supports $H_1$. Navigation is also relatively safe for the RL-only robots, i.e. collisions are low, but task time has a higher median and interquartile range in each world and there is also a consistent failure rate (17% for worlds $W_1$ and $W_2$, and 7% and 4% for worlds $W_3$ and $W_4$ respectively). The RL-only experiments thus provide some evidence to uphold $H_2$. Hypothesis $H_3$ is validated by the data in Table 4; there is both a statistical and scientific significance for the faster idiotypic task time in all of the worlds and for the lower number of collisions in the real world. However, it is worth noting that the lack of statistical collision significance in the virtual worlds may be due to the way that collisions are measured in the simulator, see Section 5. In addition, the box-and-whisker plots in Figure 9 and Figure 10 highlight a much larger spread in the data for RL-only robots for both time and collision variables. Only the real world plots are displayed, but these are typical of all the results and provide evidence to support the idiotypic system's greater reliability, as it appears to produce more consistent and predictable solutions.

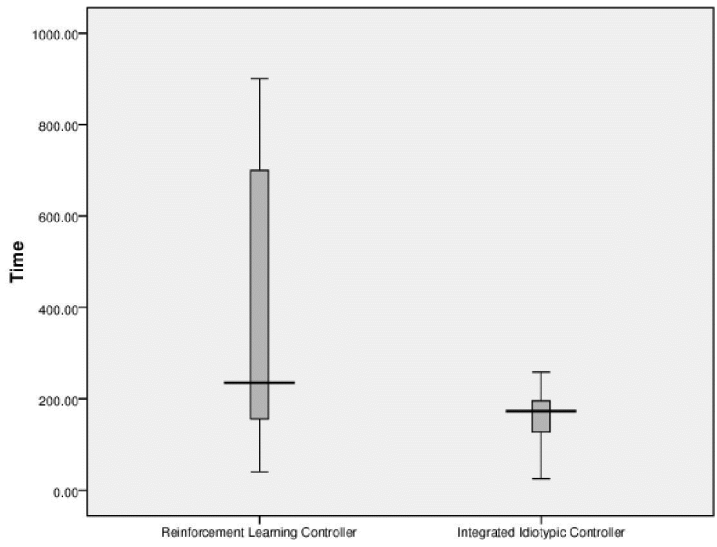

**Fig. 9** Box and whisker plot for time data in the real world

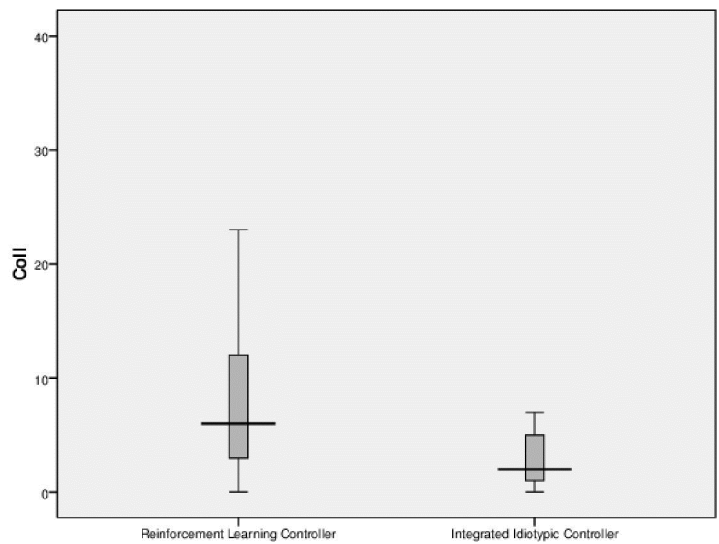

**Fig. 10** Box and whisker plot for collision data in the real world

**Table 4** Mann-Whitney and *A*−test results for differences between the idiotypic and RL-only control systems. Significant differences ($p = 0.05$) and medium and large *A*−test effects are highlighted in bold

| World | *p*-value | | *A*-value | |
|---|---|---|---|---|
| | time (s) | collisions | time (s) | collisions |
| $W_1$ | .016 | .090 | 0.68 | 0.63 |
| $W_2$ | **.000** | .067 | **0.79** | **0.64** |
| $W_3$ | **.000** | .146 | **0.77** | **0.61** |
| $W_4$ | **.005** | **.000** | **0.71** | **0.78** |

Idiotypic superiority is further illustrated in Figure 7, which shows the paths taken by a virtual, idiotypic (left) and RL-only (right) Pioneer when solving the block-finding problem in world $W_3$. It is evident that the RL-only Pioneer takes a much less direct route and repeats its path several times. This is because it is less able to adapt its behaviour and consequently spends much more time wandering, getting stuck and trying to free itself. These results are representative of the 60 runs completed in this world. Earlier work [8] suggests that the idiotypic advantage can be attributed to an increased rate of antibody change, which implies a much less greedy strategy. It also proposes that the network is capable of linking antibodies of similar type, so that useful but potentially untried ones can be used. The use of concentrations and feedback within the network may also facilitate a memory feature that achieves a good balance between selection based on past antibody use and current environmental information.

These results strongly support the notion that the idiotypic network is vitally important for adaptability and hence for achieving a successful cross-platform transfer, since it appears to provide a much more reliable methodology, and performance without it is much poorer. Another important implication of these results is the suggestion that, in principle, behaviours evolved in simulation on an epuck robot can be successfully ported to the larger-scale Pioneer P3-DX platform in both simulated and real worlds, which demonstrates that the evolutionary (LTL) phase is capable of producing sets of very diverse behaviours to aid the STL phase's idiotypic selection process.

The present scheme has some limitations; in particular, there is no scope to change the antibodies within the network, only to choose between them. A possible improvement would be constant execution of the LTL phase, which reg-

21ularly updates the genetic sequence, allowing fresh antibodies to be used as the need arises. In addition, success with transference to the Pioneer is presently too heavily dependent upon methods of inconsistency handling, in particular, readjustment of the reinforcement scheme for the particular sensor characteristics, and, more importantly, upon tuning the navigation threshold values $T_1$ and $T_2$ and the parameter $\zeta$ by initial trial and error. However, if the porting of evolved behaviours to different platforms is merely a case of making minor adjustments to a number of parameters, then one can indeed envisage a scenario where virtual, possibly fictitious, small robots may be used to evolve suitable behaviours for larger, real ones, as suggested in Section 2. Perhaps the future will see the publication of a number of look-up tables that provide a robot user with parameter values directly applicable to the model and sensors used. If this is the case, then the attainment of an adaptive control structure for a particular robot model and its sensors will have been reduced to a simple, two-step process that consists of rapid evolution with the virtual robot (to produce a number of reliable but diverse behaviour sets), followed by parameter adjustment to allow an adaptive control system to make optimum use of these behaviours.

## 7 Conclusions

This paper has identified three top-level considerations for transfer between different robotic platforms and has shown how these relate to hardware and software differences. It has also described a control architecture that consists of an LTL (evolutionary) phase responsible for the generation of diverse behaviours, and an STL (idiotypic immune system) phase, which selects from the available behaviours in an adaptive way, and has demonstrated that the behaviours are, essentially, platform independent. In particular it has shown that the behaviours can be evolved in simulation on a miniature epuck robot and used successfully on the much larger-scale Pioneer P3-DX platform in both a simulator and in the real world, provided that the idiotypic selection mechanism is switched on and that suitable scaling parameters are chosen to facilitate the desired idio-typic difference rate. Tests in a number of virtual and real-world laboratory environments have shown that the Pioneer is able to accomplish maze navigation, obstacle avoidance and target-tracking tasks successfully using the epuck behaviours, and that it performs significantly faster when employing an idiotypic mechanism, which demonstrates that idiotypic behaviour-selection is key to providing the adaptability necessary for platform transfer (equivalent to a complex and difficult environmental change). However, both the LTL and STL components are required for overall success; in the LTL phase the genetic encoding of the behaviours and the choice of separate populations that do not interbreed permits greater antibody diversity. This is important for the STL phase where idiotypic selection exploits the richness in the behaviour pool to produce a decentralized, highly-adaptive system.

**Acknowledgements** This research was funded by the Engineering and Physical Sciences Research Council (EPSRC).